\documentclass{article}

\pdfoutput=1
\usepackage{arxiv}

\usepackage[utf8]{inputenc} 
\usepackage[T1]{fontenc}    
\usepackage{hyperref}       
\usepackage{url}            
\usepackage{booktabs}       
\usepackage{amsfonts}       
\usepackage{nicefrac}       
\usepackage{microtype}      
\usepackage{lipsum}
\usepackage{fancyhdr}       
\usepackage{graphicx}       
\graphicspath{{media/}}     
\UseRawInputEncoding
\usepackage{adjustbox}
\usepackage{appendix}
\usepackage[dvipsnames]{xcolor}

\usepackage{graphicx}
\usepackage{amsmath}
\usepackage{multirow}
\usepackage{makecell}
\usepackage{bbding}

\usepackage{booktabs}
\usepackage{color}
\usepackage{colortbl}
\definecolor{ForestGreen}{RGB}{73,200,94}
\definecolor{blue1}{RGB}{25,95,225}
\definecolor{blue2}{RGB}{30,144,255}
\definecolor{blue3}{RGB}{135,206,235}
\usepackage{xspace}
\newcommand{\ourapproach}{ProtET\xspace}
\newcommand*{\affaddr}[1]{#1} 
\newcommand*{\affmark}[1][*]{\textsuperscript{#1}}

\title{Multi-Modal CLIP-Informed Protein Editing}

\begin{document}



\author{
Mingze Yin\affmark[1]\thanks{These authors contributed equally to this work.},
Hanjing Zhou\affmark[1]\footnotemark[1], 
Yiheng Zhu\affmark[1],
Miao Lin\affmark[2],
Yixuan Wu\affmark[1], 
Jialu Wu\affmark[1], \\
\textbf{Hongxia Xu\affmark[1],
Chang-Yu Hsieh\affmark[1],
Tingjun Hou\affmark[1]\thanks{Corresponding authors.},
Jintai Chen\affmark[3]\footnotemark[2],  
Jian Wu\affmark[1]\footnotemark[2]} \\
\normalsize
\affaddr{\affmark[1]Zhejiang University, Hangzhou, China} \\
\affaddr{\affmark[2]Southern Medical University, Guangzhou, China} \\
\affaddr{\affmark[3]University of Illinois at Urbana-Champaign, Urbana, USA} \\
\normalsize
{\tt \ mzyin256@gmail.com (M. Yin)}, 
{\tt \ zhj85393@gmail.com (H. Zhou)}, \\
{\tt \ jtchen721@gmail.com (J. Chen)}, \\
{\tt \ tingjunhou@zju.edu.cn (T. Hou)}, 
{\tt \ wujian2000@zju.edu.cn (J. Wu)}
}

\maketitle

\begin{abstract}
Proteins govern most biological functions essential for life, but achieving controllable protein discovery and optimization remains challenging. Recently, machine learning-assisted protein editing (MLPE) has shown promise in accelerating optimization cycles and reducing experimental workloads. However, current methods struggle with the vast combinatorial space of potential protein edits and cannot explicitly conduct protein editing using biotext instructions, limiting their interactivity with human feedback.
To fill these gaps, we propose a novel method called \ourapproach for efficient CLIP-informed protein editing through multi-modality learning. Our approach comprises two stages: in the pretraining stage, contrastive learning aligns protein-biotext representations encoded by two large language models (LLMs), respectively. Subsequently, during the protein editing stage, the fused features from editing instruction texts and original protein sequences serve as the final editing condition for generating target protein sequences. Comprehensive experiments demonstrated the superiority of \ourapproach in editing proteins to enhance human-expected functionality across multiple attribute domains, including enzyme catalytic activity, protein stability and antibody specific binding ability. And \ourapproach improves the state-of-the-art results by a large margin, leading to significant stability improvements of 16.67\% and 16.90\%. 
This capability positions \ourapproach to advance real-world artificial protein editing, potentially addressing unmet academic, industrial, and clinical needs.
\end{abstract}

\section{Introduction} 
Proteins are vital components of biological systems, executing a myriad of functions that underpin an extensive array of cellular processes and biological pathways \cite{PromptProtein}. Throughout billions of years of evolution, proteins undergo changes in their sequences and structures, which further influence their functional properties. However, accomplishing controllable protein discovery and optimization is challenging because the space of possible proteins is much larger than the space of those likely to have desired functions. Protein editing, or protein modification, is a natural process that gradually increases the diversity of protein structures and functions over time, offering valuable insights into the controlled discovery and optimization of proteins.
In the past two decades, various methodologies have been developed for post-translational protein modifications (PTMs), aiming to artificially edit amino acids to enhance human-expected properties. When applied in a biologically benign manner, these methodologies have the potential to form the foundation of true synthetic biology \cite{protein_editing_survey}, but still heavily rely on time-consuming and post-hoc wet laboratory engineering \cite{protein_editing_approach}. 

Recently, machine learning-based methods have demonstrated great promise in a wide range of protein-related applications, including 3D structure prediction \cite{alphafold-multimer, alphafold3}, mutation effects prediction \cite{deepsequence}, functionality prediction \cite{ProtTrans, xiaoshi1, xiaoshi2} and \textit{de novo} protein design \cite{progen, progen2}. With the development of Large Language Models (LLMs), Protein Language Models (PLMs) pretrained on large-scale protein sequence corpora have succeeded in acquiring powerful protein representations, showcasing outstanding performance across diverse tasks \cite{esm-1b, esm-1v, esm-2}. Researchers have also developed machine learning-assisted protein editing (MLPE) approaches, allowing \textit{in silico} searching for all edited candidates and potentially improving wet-lab protein editing performance. However, existing methods primarily apply black-box optimization algorithms to iteratively sample edited proteins and rely on fitness predictors trained on selected informative samples to guide the editing direction. The iterative refinement within the vast combinatorial space of edited protein sequences still heavily constrains the performance and efficiency of MLPE approaches. Multi-modality learning like CLIP \cite{clip} has shown promising results in image-text retrieval \cite{BLIP, BLIP2, internVL, PRIOR, FDT}, CLIP-informed image classification \cite{internVL, DL_medical_image, ImageBind, FDT}, natural language visual reasoning \cite{BLIP, triple_CL} and text-guided image editing \cite{DALLE, DALLE2, DALLE3}, illuminating an exciting opportunity on protein editing:

\emph{Can we leverage multi-modality aligned protein language models to efficiently generate edited proteins under the guidance of biological texts?}

\begin{figure*}[t]
    \centering
    \includegraphics[width=0.90\textwidth]{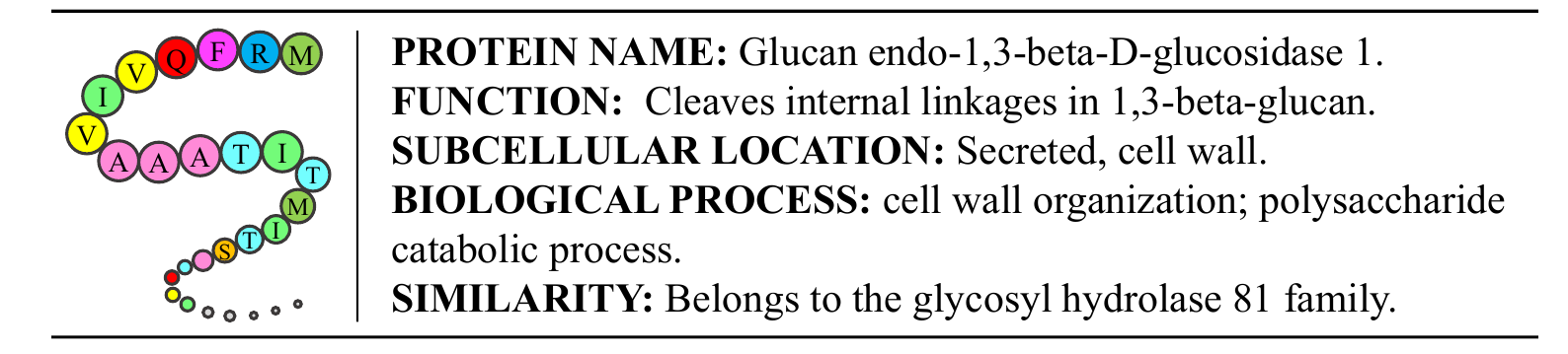}
    \caption{An illustration of the protein-biotext pair. The textual descriptions include the protein's name, function, subcellular location, biological process, and similarity to other proteins.}
    \label{fig1}
\end{figure*}

To this end, we propose a generic protein editing method named \ourapproach. \ourapproach is a two-stage CLIP-informed multi-modal approach that can accomplish proximally constrained protein editing towards desired properties. Specifically, we first curate millions of protein-biotext aligned pairs, each comprising protein sequences and functional biotext annotations, as illustrated in Figure~\ref{fig1}. We then construct transformer-structured encoder-based models (\textit{i.e.}, a large protein model and a large language model) to encode the features of both protein sequences and biotexts, respectively. Similar to CLIP \cite{clip}, multi-modality pretraining is performed using contrastive learning objectives to align the features of the protein and biotext, facilitating easier editing instruction. In the editing stage, the aligned protein features and desired function description features extracted by the pretrained models are fused into a decoder model to generate the desired protein sequences in an auto-regressive manner.

To comprehensively assess the capability of \ourapproach, we conduct holistic experiments to evaluate its superiority in editing proteins with improved human-expected functions. 
\ourapproach achieves state-of-the-art performance in protein function classification tasks, verifying the effectiveness of multi-modal protein-biotext pretraining informed by CLIP \cite{clip}. Compared to existing machine learning based protein editing methods (Single-Mutant \cite{single_mutant}, AFP-DE \cite{AFP-DE}, and EvoPlay \cite{EvoPlay}), \ourapproach demonstrates promising performance on the established protein editing task, achieving better editing performance that improves the stability of original protein sequences by 16.67\% and 16.90\% under two different stability assessment criteria. Additionally, the 
enzymes designed by \ourapproach realize a significant leap in the catalytic activity. Antibodies optimized by \ourapproach also form stable, regular 3D structures binding with SARS-CoV-1 and SARS-CoV-2 antigens.
These experimental results highlight \ourapproach as a valuable tool for future controllable protein discovery and optimization endeavors in real-world scenarios.

\section{Methods}

\begin{figure}[h]
    \centering
    \begin{minipage}{0.45\textwidth}
        \includegraphics[width=0.98\textwidth]{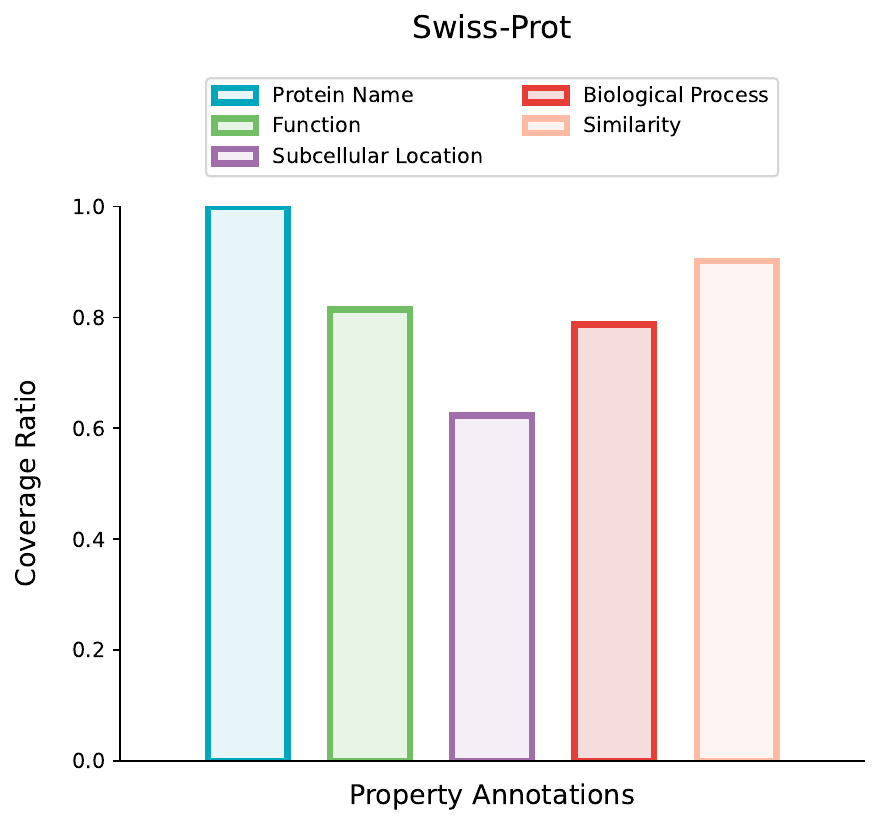}
        \label{fig2a}
    \end{minipage}
    \begin{minipage}{0.45\textwidth}
        \includegraphics[width=0.98\textwidth]{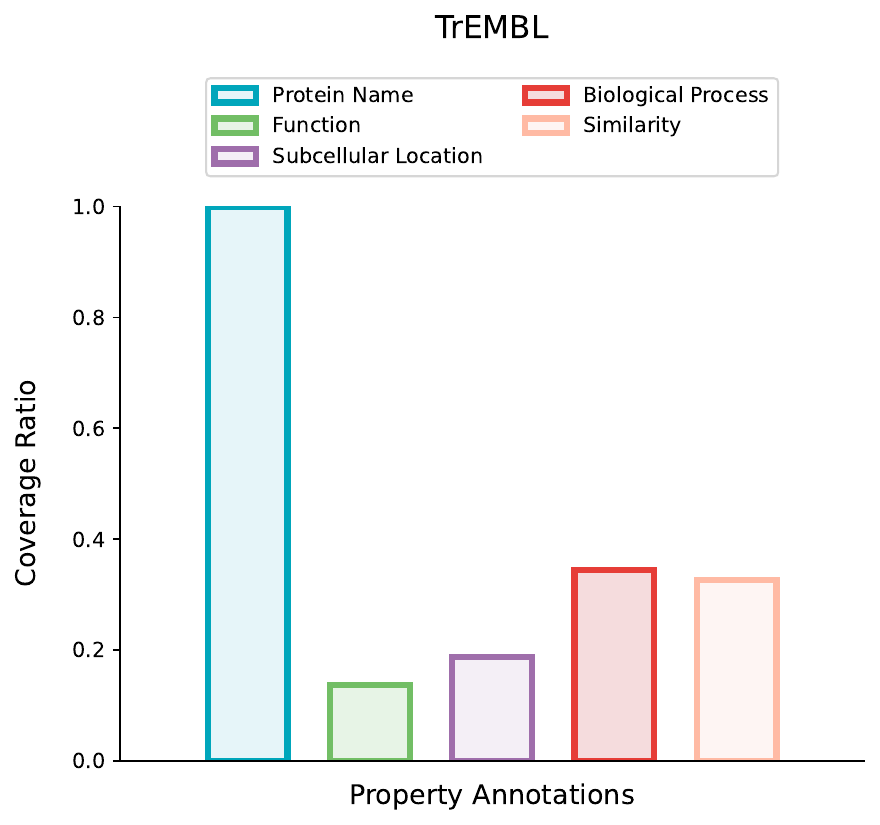}
        \label{fig2b}
    \end{minipage}
    \caption{Coverage ratios of protein property annotations in Swiss-Prot and TrEMBL.}   
    \label{fig2}
\end{figure}

\begin{table}[b]
    \caption{Protein sequence statistics of different evidence level in Swiss-Prot and TrEMBL.}    
    \centering
    \setlength{\tabcolsep}{0.2em}
    \resizebox{1.0\textwidth}{!}{\begin{tabular}{l|ccccc}
            \hline
            \multicolumn{6}{c}{Swiss-Prot} \\
            \hline
            evidence level $\downarrow$ & ``Protein Name'' & ``Function'' & ``Subcellular Location'' & ``Biological Process'' & ``Similarity'' \\
            \hline
            1: Evidence at protein level & 113,077 & 100,669 & 85,666 & 92,319 & 90,599 \\
            2: Evidence at transcript level & 55,883 & 45,001 & 44,158 & 41,718 & 45,961 \\
            3: Inferred from homology & 386,595 & 317,164 & 222,312 & 312,668 & 376,527 \\
            4: Predicted & 13,032 & 1,600 & 3,048 & 2,282 & 810 \\
            5: Uncertain & 1,833 & 303 & 732 & 483 & 728 \\
            \hline
            \multicolumn{6}{c}{TrEMBL} \\
            \hline
            evidence level $\downarrow$ & ``Protein Name'' & ``Function'' & ``Subcellular Location'' & ``Biological Process'' & ``Similarity'' \\
            \hline
            1: Evidence at protein level & 230,045 & 24,700 & 57,303 & 113,846 & 96,717 \\
            2: Evidence at transcript level & 1,373,855 & 265,605 & 417,404 & 637,893 & 687,615 \\
            3: Inferred from homology & 81,928,782 & 32,589,671 & 35,259,878 & 54,799,521 & 81,090,429 \\
            4: Predicted & 167,598,957 & 1,573,895 & 11,568,640 & 30,878,461 & 31,248 \\
            5: Uncertain & 0 & 0 & 0 & 0 & 0 \\
            \hline
            \end{tabular}
            }
    \label{tab1}
\end{table}


\subsection{Curated protein-biotext dataset}
To attain multi-modal protein-biotext pretraining, we first build a protein-biotext paired dataset. 
Swiss-Prot database strives to provide a high level of manually reviewed protein annotations, with a minimal level of redundancy \cite{protein_data}. And TrEMBL consists of protein entries with computationally analyzed annotations, derived from the translation of EMBL nucleotide sequences \cite{protein_data}.
Given protein database with rich, consistent and accurate protein functional annotations, we curate a new multi-modal dataset with aligned pairs of protein sequences and biotext functional annotations \cite{Uniprot}. 
Concretely, proteins with elaborate annotations are downloaded from Swiss-Prot and TrEMBL in Jan. 2024, yielding 570,420 and 251,131,639 sequences respectively.
for protein sequences collected from Swiss-Prot and TrEMBL, we select five property fields: 
(1) \textit{"Protein Name"} gives the full protein name recommended by the UniProt consortium; 
(2) \textit{"Function"} depicts diverse functions owned by a protein; 
(3) \textit{"Subcellular Location"} describes the location and topology of a mature protein in the cell; 
(4) \textit{"Biological Process"} represents larger processes accomplished by multiple molecular activities;
(5) \textit{"Similarity"} provides information about the protein families that a protein belongs to.
The above biotext descriptions and the corresponding protein sequences are aligned in detail, as shown in Figure~\ref{fig1}. 
Additionally, we record the annotation coverage ratio of the aforementioned five selected property fields, as displayed in Figure~\ref{fig2}. For existing property annotations, we present the detailed protein sequence amount at different evidence levels in Table~\ref{tab1}.
To further improve the quality of curated multi-modal dataset, we keep all manually reviewed protein annotations from Swiss-Prot and meticulously filter computationally analyzed protein annotations from TrEMBL. 
Specifically, we remove the protein entries having annotation coverage ratio less than 40\%, as well as those with low evidence level at 4 and 5, resulting in 67,972,109 protein-biotext aligned pairs tailored for multi-modality pretraining. 

\begin{figure*}[!ht]
    \centering
    \includegraphics[width=0.95\textwidth]{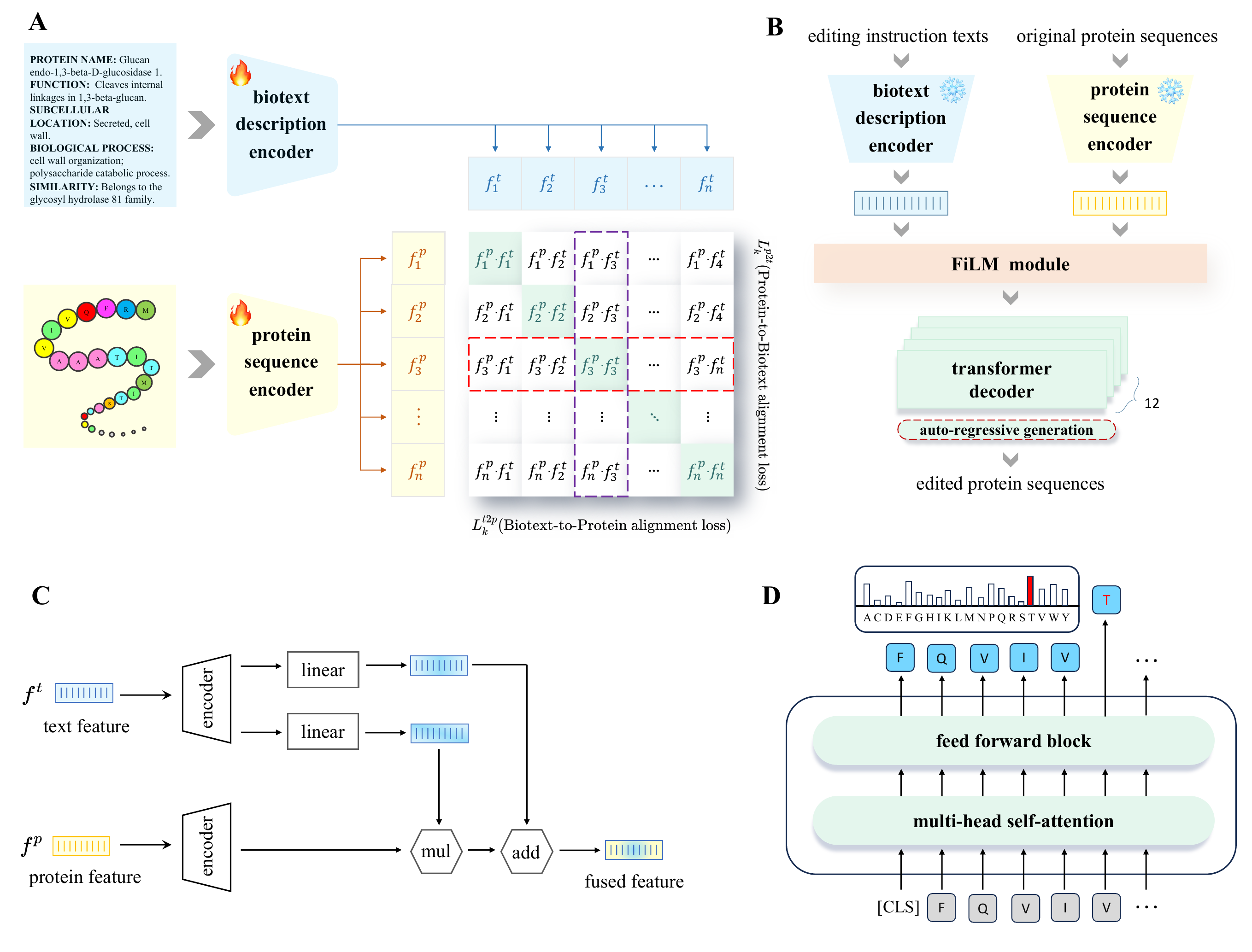}
    \caption{
    The workflow and framework details of \ourapproach.
    (A) A CLIP-like contrastive pretraining aligns features of protein sequences and biotext descriptions.
    (B) The FiLM module and transformer-decoders for protein editing. The FiLM module integrates multimodal features from the original protein sequences and the editing instruction texts, serving as the editing condition. Based on this condition, transformer-decoders design edited protein sequences through an autoregressive generation process. 
    (C) Details of the FiLM module. 
    It extracts multiplicative and additive factors from text features using linear mappings. These factors conditionally optimize protein features through addition and multiplication, to create fused features.
    (D) Details of the Transformer decoder. It uses a multi-head self-attention module to learn comprehensive residue interactions and predicts the next residue based on the previous ones.}
    \label{fig3}
\end{figure*}

\subsection{Multi-modality pretraining}
\label{sec:pretraining}
Aiming to pave the way for CLIP-informed protein editing, we first need to extract features of the protein and biotext, and align feature spaces of both modalities. 
In this paper, we exploit two transformer-encoder based large language models to extract protein and biotext features respectively. 
ESM-2 is employed as the protein sequence encoder, which is pretrained on millions of protein sequences \cite{esm-2}. In recent studies, ESM-2 has proven quite beneficial in many protein-related studies~\cite{ProtST, OntoProtein}.
PubMedBERT is employed as the biotext description encoder, which is pretrained on PubMed articles context \cite{PubMedBERT}. PubMedBERT is a robust biomedical language model customized to encode biomedical texts.

Inspired by extraordinary endeavors of image-text multi-modality alignment, we propose to align feature spaces of both modalities via contrastive learning \cite{clip, DALLE, DALLE2, groupon}. 
As illustrated in Figure~\ref{fig3}A, in a list of protein-biotext pairs, $p_i$ and $t_j$ represent the $i_{th}$ protein sequence and $j_{th}$ biotext description, respectively. When $i = j$, $(p_i,t_j)$ is a positive pair, and when $i \neq j$ it is a negative pair. $f_i^p$ and $f_i^t$ are protein and biotext representations extracted by the aforementioned encoders, both projected to the same feature dimension. Formally, we introduce our contrastive learning objective, training our model to discriminate positive and negative protein-biotext pairs. Given a batch of N protein-biotext pairs, the overall training objective is constructed of the Protein-to-Biotext alignment loss and Biotext-to-Protein alignment loss:
\begin{equation}
    \mathcal{L}_i^{p2t}=-\frac{1}{N}\log\frac{\exp(\mathtt{sim}(f_i^p,f_i^t)/\tau)}{\sum_j\exp(\mathtt{sim}(f_i^p,f_j^t)/\tau)},
\end{equation}
\begin{equation}
    \mathcal{L}_i^{t2p}=-\frac{1}{N}\log\frac{\exp(\mathtt{sim}(f_i^t,f_i^p)/\tau)}{\sum_j\exp(\mathtt{sim}(f_i^t,f_j^p)/\tau)},
\end{equation}
where $\mathtt{sim}(\cdot,\cdot)$ denotes the similarity function (\textit{i.e.}, vector dot product in this paper) and $\tau$ denotes the temperature parameter that controls the softmax distribution. 

Overall, the final contrastive learning objective in the multi-modality pretraining stage is formulated as:
\begin{equation}
    \mathcal{L}_{align}=\frac{1}{2}\sum_{i=1}^N(\mathcal{L}_i^{p2t}+\mathcal{L}_i^{t2p}).
\end{equation}
where $\mathcal{L}_i^{p2t}, \mathcal{L}_i^{t2p}$ respectively represent the Protein-to-Biotext alignment and Biotext-to-Protein alignment loss.

\subsection{Protein editing generator}
\label{sec:generation}
Owing to the aligned feature spaces accomplished by multi-modality pretraining introduced in Section~\ref{sec:pretraining}, now we could encode the features of original protein sequences and editing instruction texts, then construct a decoder to generate the edited protein sequences.
As shown in Figure~\ref{fig3}B, the protein editing generator is composed of a feature fusion module and a generative decoder.
Concretely, FiLM module \cite{FiLM} is leveraged to fuse multi-modal features from original protein sequences and editing instruction texts. Such architecture design aims to optimize protein features based on editing instruction text features, integrating multi-modal information to accomplish cross-modal protein editing. 
The generative decoder is intentionally constructed using 12 layers transformer decoders, tailored for auto-regressively generating edited protein sequences. Given fused features as the final editing condition, edited protein sequences are generated incrementally, one amino acid at a time, with each amino acid being conditioned on the probability of previously generated ones.
\begin{equation}
P(S_{edited}) = \prod_{i=1}^{n} P(s_i | s_1, s_2, ..., s_{i-1}; \texttt{FiLM}(S_{original}, T_{instruction}))
\end{equation}
where $S_{edited}=(s_1, s_2, ..., s_n)$ represents the edited protein sequences and $(S_{original}, T_{instruction})$ represents encoded features of original protein sequences and editing instruction texts, respectively. 
Given the paired original and edited protein sequences, the protein editing generator is trained in an unsupervised manner to generate proteins with higher feature similarity to the editing instruction texts, aligning with human-expected functional attributes.
\begin{equation}
\begin{aligned}
\mathcal{L}_{edit} = \mathtt{H}\left[\mathtt{sim}(S_{original},T_{instruction}) - \mathtt{sim}(S_{edited},T_{instruction})\right] \\
+ \hspace{0.2cm} \texttt{SoftCrossEntropy}\left[ y(s_i/S_{original}), p(s_i/S_{edited}) \right]
\end{aligned}
\end{equation}
where $s_i/S_{original}$ and $s_i/S_{edited}$ respectively denote the amino acid in the \textit{ith} position of original protein sequences and edited protein sequences. $\mathtt{sim}(\cdot,\cdot)$ denotes the cosine similarity function and $\mathtt{H}(\alpha) = max(0, \alpha)$. The second item in $\mathcal{L}_{edit}$ serves as a regularization term to avoid model collapse and guides the model to perform slight amino acid modification, gradually improving functional attributes.

\subsection{Implementation details}
Following previous works \cite{OntoProtein, ProtST, BioBridge}, we use the base-version of ESM-2 model with 650 million trainable parameters to encode protein sequences \cite{esm-2}. As for biotext encoding, we apply PubMedBERT with 12 transformer layers and 100 million trainable parameters. 
All protein sequences are padded or truncated to a fixed length of 1024 tokens, and we align all the biotext descriptions to a unified length of 512. The embedding dimensions of protein and biotext LLMs are 1280 and 768 respectively. Aiming to align multi-modal representations, we choose 512 to be the projected common feature dimension. The temperature coefficient $\tau$ is set as 0.01. 
For the curated multi-modal dataset, we use protein-biotext pairs from filtered TrEMBL for the pretraining stage, and those from Swiss-Prot are exploited to train protein editing generator.
The overall framework are trained with a batch size of 128 for 10 epochs, utilizing 16 NVIDIA 32G V100 GPUs. The learning rate is initialized as $5e^{-5}$ with 2,000 linear warm-up steps.

\section{Results}
\subsection{Protein function classification}
\subsubsection{Problem setup}
Following CLIP \cite{clip}, our proposed method incorporates a large-scale multi-modal pretraining stage, before executing the cross-modal protein editing. We first conduct experiments to validate that the PLM's functional protein understanding is enhanced by multi-modality learning with the biotext. Specifically, four standard protein function classification benchmarks curated by DeepFRI \cite{DeepFRI} are employed to classify proteins with multiple functional labels, including Enzyme Commission (EC), Gene Ontology Biological Process (GO-BP), Gene Ontology Molecular Function (GO-MF) and Gene Ontology Cellular Component (GO-CC).
Following the previous work~\cite{ProtST}, we exploit the dataset splits under 95\% sequence identity cutoff for both EC and GO. And we consider four traditional models (CNN, ResNet, LSTM and Transformer) and three unimodal pretrained PLMs (ProtBERT~\cite{ProtTrans}, ESM-1b~\cite{esm-1b}, ESM-2~\cite{esm-2}) as baselines. Furthermore, we include two PLMs (OntoProtein~\cite{OntoProtein}, ProtST-ESM2~\cite{ProtST}) undergoing multi-modal alignment with the biotext to further evaluate the superiority of our multi-modal pretraining stage. And function classification results are measured by AUPR and F$_{\text{max}}$.

\begin{table}[ht]
    \caption{Results of the protein function classification tasks. We highlight the best results in \textbf{bold} and \underline{underline} the second best results.}
    \vspace{-1em}
    \label{tab2}
    \begin{center}
    \resizebox{\textwidth}{!}{
    \begin{tabular}{c|l|c c| c c| c c| c c}
            \toprule
            \multirow{2}{*}{\textbf{Setting}} & \multicolumn{1}{c|}{\multirow{2}{*}{\textbf{Method}}} & \multicolumn{2}{c|}{\textbf{EC}} & \multicolumn{2}{c|}{\textbf{GO-BP}} & \multicolumn{2}{c|}{\textbf{GO-MF}} & \multicolumn{2}{c}{\textbf{GO-CC}}\\
            \cmidrule{3-10}
            & & \small{AUPR} & $F_{\text{max}}$ & \small{AUPR} & $F_{\text{max}}$ & \small{AUPR} & $F_{\text{max}}$ & \small{AUPR} & $F_{\text{max}}$ \\
            \midrule
            \multirow{4}{*}{\makecell{Traditional \\ models}} & CNN & 0.540 & 0.545 & 0.165 & 0.244 & 0.380 & 0.354 & 0.261 & 0.387 \\
            & ResNet & 0.137 & 0.187 & 0.166 & 0.280 & 0.281 & 0.267 & 0.266 & 0.403 \\
            & LSTM & 0.032 & 0.082 & 0.130 & 0.248 & 0.100 & 0.166 & 0.150 & 0.320 \\
            & Transformer & 0.187 & 0.219 & 0.135 & 0.257 & 0.172 & 0.240 & 0.170 & 0.380 \\
            \midrule
            \multirow{6}{*}{\makecell{PLMs under \\ full fine-tuning}} & 
            ProtBERT & 0.859 & 0.838 & 0.188 & 0.279 & 0.464 & 0.456 & 0.234 & 0.408 \\
            & OntoProtein & 0.854 & 0.841 & 0.284 & 0.436 & 0.603 & 0.631 & 0.300 & 0.441 \\
            & ESM-1b & 0.884 & 0.869 & 0.332 & 0.452 & 0.630 & 0.659 & 0.324 & 0.477 \\
            & ESM-2 & 0.888 & 0.874 & 0.340 & 0.472 & 0.643 & 0.662 & 0.350 & 0.472 \\
            & ProtST-ESM2 & \underline{0.898} & \underline{0.878} & \underline{0.342} & \underline{0.482} & \underline{0.647} & \underline{0.668} & \textbf{0.364} & \textbf{0.487} \\
            & \ourapproach & \cellcolor{gray!20}\textbf{0.901} & \cellcolor{gray!20}\textbf{0.883} & \cellcolor{gray!20}\textbf{0.351} & 
            \cellcolor{gray!20}\textbf{0.489} & 
            \cellcolor{gray!20}\textbf{0.649} & 
            \cellcolor{gray!20}\textbf{0.673} & 
            \cellcolor{gray!20}\underline{0.362} &
            \cellcolor{gray!20}\underline{0.486} \\
            \bottomrule
    \end{tabular}}
    \end{center}
\end{table}

\subsubsection{Experimental results}
As shown in Table~\ref{tab2}, \ourapproach achieves state-of-the-art performance on 6 out of 8 evaluation metrics. And we observe that \ourapproach clearly outperforms the vanilla unimodal PLMs and previous multi-modal aligned PLMs, while the large-scale PLMs performs consistently better than traditional models. These results demonstrate that our multi-modal protein-biotext pretraining process is generally beneficial to protein functional understanding, which boosts performance on diverse classification tasks.

\subsection{Enzyme catalytic activity editing}
\subsubsection{Problem setup}
\begin{figure*}[!ht]
    \vspace{0.3cm}
    \centering
    \includegraphics[width=0.80\textwidth]{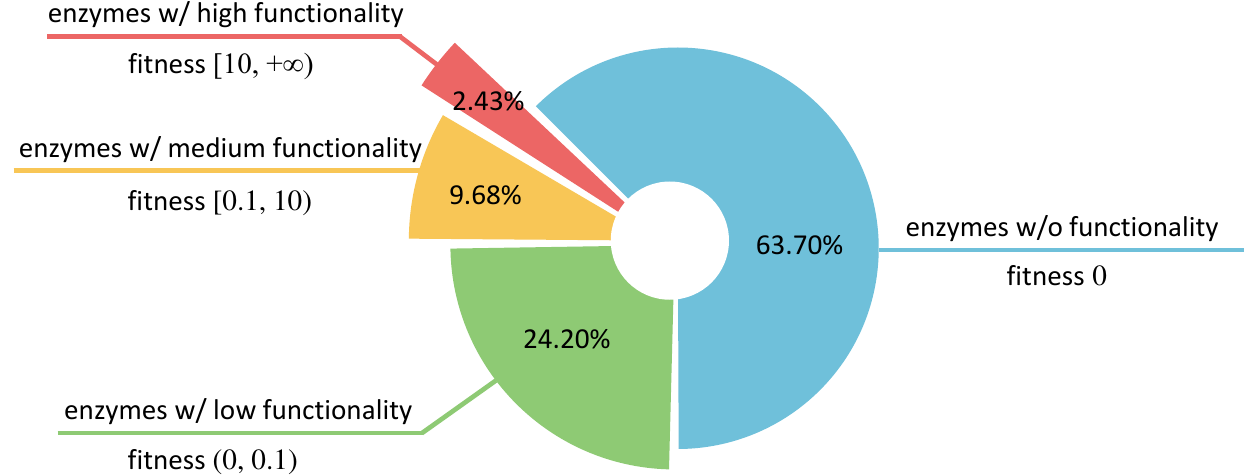}
    \vspace{1em}
    \caption{The compositional structure of the enzyme dataset. According to annotated catalytic activity scores, the enzyme dataset is divided into four subsets with different fitness levels, and we present the proportion of data for these constructed subsets.}
    \label{fig4}
\end{figure*}

Enzymes are vital proteins that promote metabolism, or the chemical reactions in biological activities. Thus we conduct a protein editing experiment on publicly available PhoQ \cite{PhoQ} dataset, aiming to optimize enzymes towards higher catalytic activity. PhoQ is one of the most widely used dataset to test the protein editing capability of MLPE models. It consists of 140,517 enzymes at four sites (A284, V285, S288 and T289), annotated with catalytic activity scores. And the catalytic activity scores in PhoQ represents the phosphatase or kinase activity of different PhoQ mutants. Note that we only include the enzymes with thoroughly annotated catalytic activity scores in this task. Given diverse enzymes with annotated catalytic activity scores, we divide the enzyme dataset into subsets with \textit{high}, \textit{medium}, and \textit{low} functionality, as well as a subset without functionality (\textit{i.e.}, enzyme catalytic activity, also described as fitness). The compositional structure of the enzyme dataset with different annotated fitness levels are illustrated in Figure~\ref{fig4}. We respectively sample 100 enzymes from subsets with different fitness levels for the protein editing test and visualization, leaving the remaining proteins for model fine-tuning, and only the aligned key sites are included for loss computation.

\subsubsection{Experimental results}
\begin{figure}[ht]
    \centering
    \begin{minipage}{0.48\textwidth}
        \includegraphics[width=0.98\textwidth]{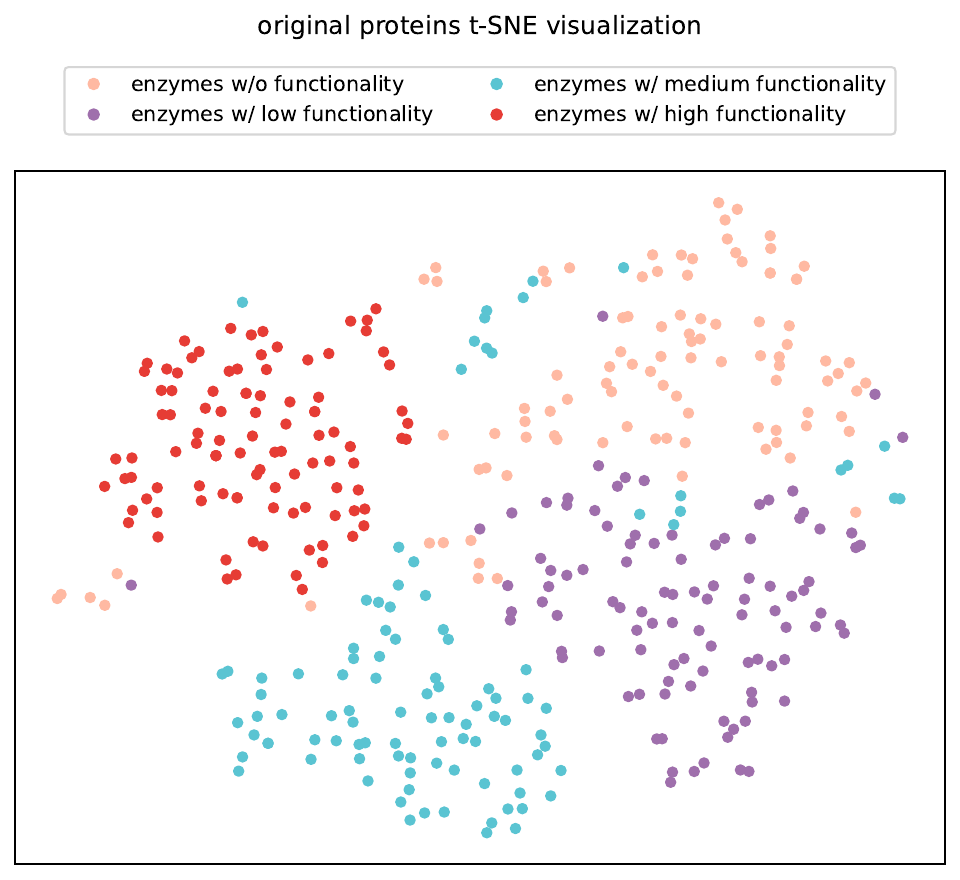}
        \label{fig5a}
    \end{minipage}
    \begin{minipage}{0.48\textwidth}
        \includegraphics[width=0.98\textwidth]{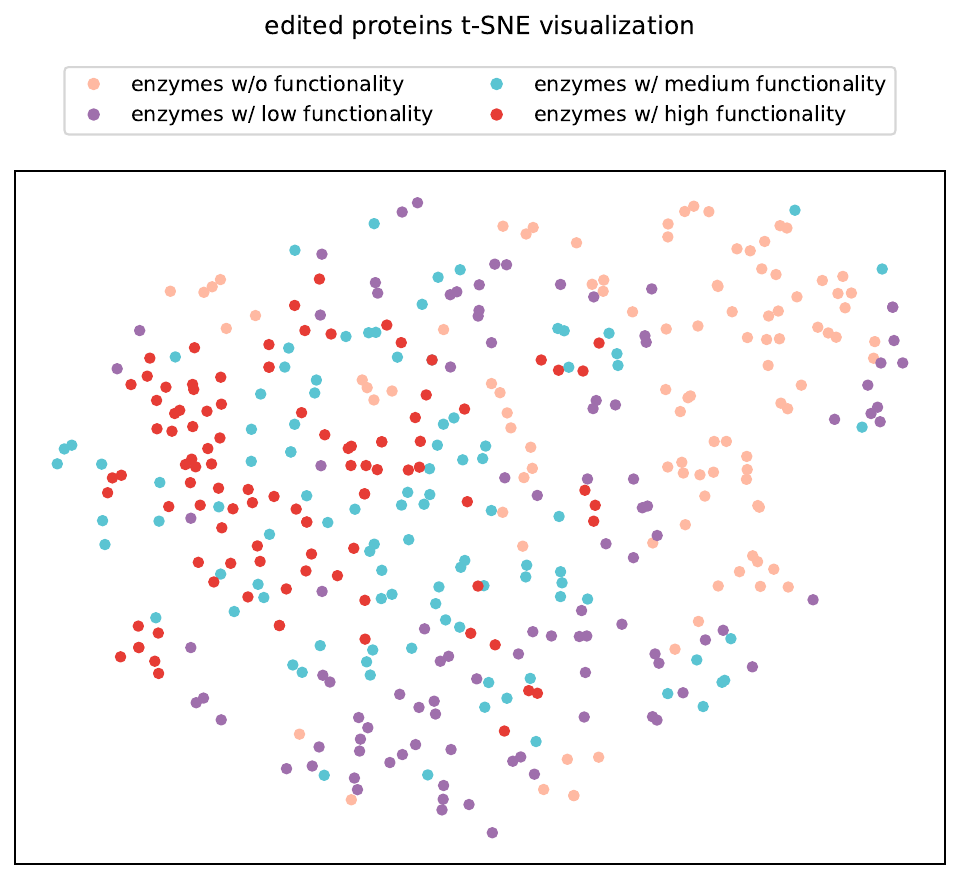}
        \label{fig5b}
    \end{minipage}
    \caption{The t-SNE visualization results. Different colors indicate enzymes with different fitness levels correspondingly. Enzymes with \textit{medium}, \textit{low} and \textit{zero} fitness tend to cluster together with \textit{high}-functionality enzymes after being edited by \ourapproach.}   
    \label{fig5}
\end{figure}

We display t-SNE visualization results in Figure~\ref{fig5}. 
First, we observe that original enzymes with different fitness levels are nicely clustered together, as shown in Figure~\ref{fig5}(left), indicating the catalytic activity information can serve as an important clustering basis.
Then we exploit \ourapproach to perform CLIP-informed protein editing on three enzyme subsets with \textit{medium}, \textit{low} and \textit{zero} fitness, leaving the enzyme subset with \textit{high} fitness as the editing reference (\textit{i.e.}, enzymes with \textit{high} fitness serve as the golden standard in this experiment). 
Owing to the powerful editing capability of \ourapproach, three enzyme subsets initially with poor functionality are optimized and move closer to the editing reference (\textit{i.e.}, enzyme subset with \textit{high} fitness), as shown in Figure~\ref{fig5}(right). Such phenomenon manifests that enzymes edited and optimized by \ourapproach fulfill a significant leap in the catalytic activity.

\subsection{Protein stability editing}
\label{protein_stability_editing}
\subsubsection{Problem setup}
To better evaluate the protein editing capability of \ourapproach, it is worthwhile to conduct convincing experiments. 
Designing stable proteins is important to ensure, for instance, that drugs are delivered before they are degraded. More generally, given a broad sample of protein measurements, finding better refinements of top stable candidates is useful for maximizing yield from expensive protein engineering experiments. 
We evaluate \ourapproach on a set of protein sequences with stability annotations curated by~\cite{TAPE}. Starting from original protein sequences, we aim for the edited proteins to exhibit better stability, maintaining its fold above a concentration threshold.
As for comprehensive comparison, the stability scores of original protein sequences are included as the bottom line of protein editing performance. We also adopt a single mutation walk approach, referred to as Single-Mutant, as the protein editing baseline~\cite{single_mutant}, which is a hill-climbing algorithm. Single-Mutant~\cite{single_mutant} performs the single-site mutation across the full-length of original protein sequences, and select the mutated protein sequences with the highest stability score as the final edited protein sequences. Furthermore, we compare \ourapproach with other deep learning based methods, including AFP-DE~\cite{AFP-DE} and EvoPlay~\cite{EvoPlay}. Concretely, AFP-DE~\cite{AFP-DE} exploits the actively fine-tuned protein language model as the sampler and identifies informative mutants that are both representative and diverse. EvoPlay~\cite{EvoPlay} mutates a single-site residue as an action to optimize protein sequences, likening the protein optimization process to playing pieces on a chessboard. 

As for the measurement of protein stability, we employ two computational approaches to assess the stability scores of edited protein sequences. The simpler one is to directly compute the cosine similarity between the representations of editing instruction texts and edited protein sequences. For the more complex one, we train a Multi-Layer Perceptron (MLP) according to the experimental evaluated protein stability labels with a regression loss, and exploit it as the oracle (\textit{i.e.}, surrogate of biological stability experiments) to score stability of edited protein sequences. 

\subsubsection{Experimental results}
\begin{table}[htbp]
\begin{minipage}{0.48\textwidth}
\centering
\caption{Results on the stability score of edited proteins. We record \textbf{stability scores calculated by cosine similarity} here. The \textcolor{blue2}{best} and \textcolor{blue3}{second-best} results are marked.
}
\label{tab3}
\begin{small}
\begin{tabular}{c|c}
\toprule
\textbf{Method} & \textbf{stability score} \\
\midrule
original protein sequences & 0.54 \\
\midrule
Single-Mutant & 0.56 \\
\midrule
AFP-DE~\cite{AFP-DE} & \cellcolor{blue3}0.61 \\
\midrule
EvoPlay~\cite{EvoPlay} & 0.59 \\
\midrule
\ourapproach & \cellcolor{blue2}0.63 \\
\bottomrule
\end{tabular}
\end{small}
\end{minipage}
\hfill
\begin{minipage}{0.48\textwidth}
\centering
\caption{Results on the stability score of edited proteins. We record \textbf{stability scores calculated by the oracle} here. The \textcolor{blue2}{best} and \textcolor{blue3}{second-best} results are marked.
}
\label{tab4}
\begin{small}
\begin{tabular}{c|c}
\toprule
\textbf{Method} & \textbf{stability score} \\
\midrule
original protein sequences & 0.71 \\
\midrule
Single-Mutant & 0.70 \\
\midrule
AFP-DE~\cite{AFP-DE} & 0.77 \\
\midrule
EvoPlay~\cite{EvoPlay} & \cellcolor{blue3}0.82 \\
\midrule
\ourapproach & \cellcolor{blue2}0.83 \\
\bottomrule
\end{tabular}
\end{small}
\end{minipage}
\end{table}
\begin{figure}[!ht]
    \centering
    \includegraphics[width=0.7\columnwidth]{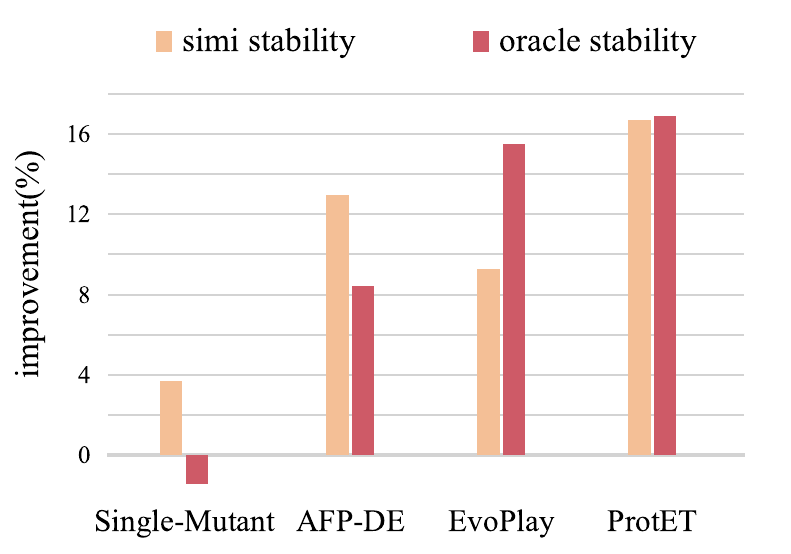}
    \vspace{-0.5em}
    \caption{Results on stability improvement of edited proteins. simi stability: the improvement percentage of stability measured by cosine similarity. oracle stability: the improvement percentage of stability measured by the conducted oracle.}
    \label{fig6}
\end{figure}
As shown in Table~\ref{tab3} and Table~\ref{tab4}, we find that \ourapproach generates the edited protein sequences with the highest stability under both of the aforementioned assessment criteria. For the first criterion (cosine similarity), \ourapproach achieves state-of-the-art performance although other baselines also manifest effective improvement. For the second criterion (oracle), EvoPlay~\cite{EvoPlay} demonstrates considerable improvements over all other baselines, whereas \ourapproach slightly outperforms EvoPlay.

Furthermore, Figure~\ref{fig6} displays the stability improvements of edited protein sequences compared to the corresponding original protein sequences. We observe significant stability enhancements in 16.67\% and 16.90\% of the proteins edited by \ourapproach, which adequately illustrates the superiority of the proposed protein editing method.

\subsection{Zero-shot SARS-CoV antibody optimization}
\subsubsection{Problem setup}
The versatile binding properties of antibodies have made them a significantly important category of proteins \cite{EATLM, IgLM, AbLang, AbLang2}. Among antibodies, Complementarity Determining Regions (CDRs) are the key component to determine the specificity and binding affinity, while CDR-H3 exhibits the highest degree of variability and is hard to predict~\cite{ABGNN}. Therefore, we further evaluate \ourapproach to optimize specific binding affinities of SARS-CoV antibodies in a zero-shot manner. We randomly sample 100 antibodies binding to SARS-CoV-1 or SARS-CoV-2 from CoV-AbDab~\cite{CoV-AbDab}. Note that \ourapproach focuses exclusively on optimizing CDR-H3 of SARS-CoV antibodies in this task.

Given antibody sequences binding to specific antigens (\textit{i.e.}, SARS-CoV-1 or SARS-CoV-2), we perturb antibody sequences by randomly substituting amino acids in the CDR-H3 region with a 15\% probability. Then \ourapproach is exploited to edit antibody CDR-H3 fragments with noise, aiming to obtain stronger antigen-antibody binding affinity. Specifically, we employ the pretrained model to generate numerous optimized CDR-H3 fragments, without further fine-tuning on antibody specific data. For each antibody CDR-H3 fragments with noise, 100 CDR-H3 samples are generated by our \ourapproach framework. Then the generated CDR-H3 are combined with standard framework regions, enabling 100 full-length generated antibody heavy chains. 
Since naturalness has been widely proven to be one of the effective indicators reflecting the potential functionality of protein sequences~\cite{progen, IgT5}, we select the $top 5$ antibody heavy chains with the highest naturalness scores using ProGen2~\cite{progen2}. 
The optimized antibody sequences are evaluated by whether they possess the capability to fold into regular 3D structures.

\subsubsection{Experimental results}
\begin{figure}[!ht]
    \centering
    \includegraphics[width=1.0\textwidth]{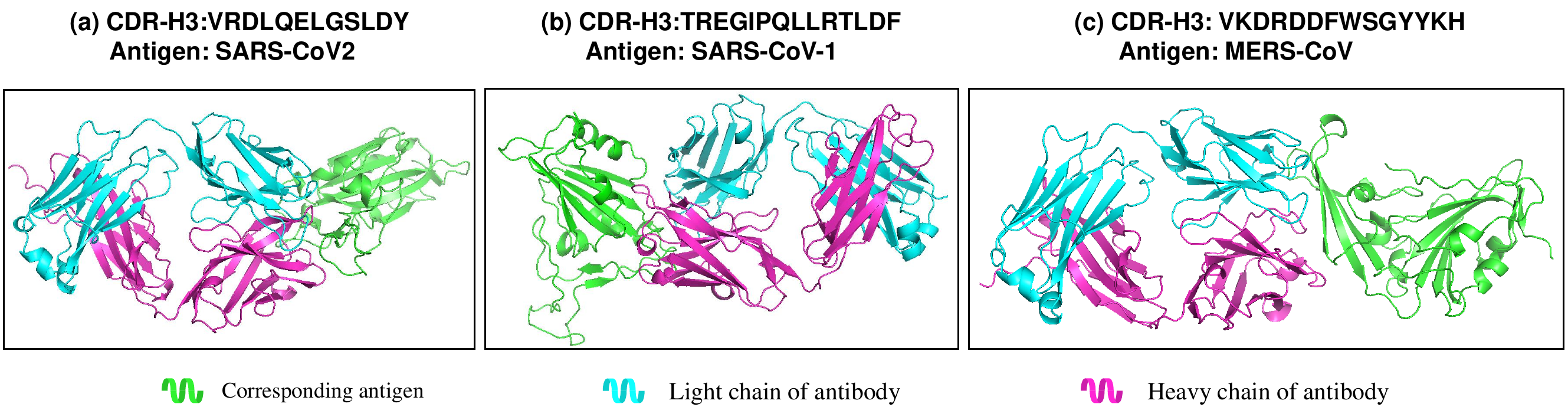}
    \caption{Alphafold-estimated 3D structures of complexes comprising optimized antibodies and corresponding antigens.}
    \label{fig7}
\end{figure}

To measure the foldability of the designed antibody sequences, Alphafold-Multimer \cite{alphafold-multimer} are exploited to predict 3D structures comprising the antigen and designed antibodies (including light chains and selected heavy chains).
We visualize the generated 3D structures with the highest Alphafold pLDDT.
As illustrated in Figure~\ref{fig7}, optimized antibodies are found to form stable, regular 3D structures binding with SARS-CoV-1 or SARS-CoV-2 antigens, manifesting our model successfully designs proteins that satisfy structural constraints.

\subsection{Ablation study}
In this section, we execute the comprehensive ablation study of \ourapproach using the protein stability editing dataset described in Section~\ref{protein_stability_editing}. To thoroughly examine the effectiveness of our proposed innovations, the ablation study focus on following perspectives: 
(1) \textbf{constructed protein-biotext paired dataset}, uses annotation coverage and evidence level metrics to filter protein entries from Swiss-Prot and TrEMBL, constructing the high-quality protein-biotext dataset to promote the multi-modality learning.
(2) \textbf{multi-modal pretraining stage}, which is constructed to align the feature spaces of protein and biotext, informed by CLIP \cite{clip}, lays the groundwork for cross-modal protein editing.
(3) \textbf{the FiLM module}, is applied for multi-modal feature fusion, with the fused feature acting as the final condition for cross-modal protein editing. And we evaluate the essentiality of the FiLM module by replacing it with a simple concatenation operation.
As presented in Table~\ref{tab5}, we notice that all proposed innovations play a significant role in the model's performance. In particular, the absence of multi-modal pretraining stage yields the highest decrease in performance. The holistic ablation study further validates the superiority of the proposed protein editing method.

\begin{table}[htbp]
\caption{Ablation results of the proposed method on the protein stability editing experiment. \textbf{CPD} denotes the constructed protein-biotext paired dataset, \textbf{MMP} denotes the multi-modal pretraining stage and \textbf{FiLM} represents the FiLM module for the feature fusion.}
\label{tab5}
\begin{center}
\begin{small}
\begin{tabular}{ccc|c|c}
\toprule
\multicolumn{3}{c|}{Components} & \multirow{2}{*}{simi stability} & \multirow{2}{*}{oracle stability} \\
\cmidrule{1-3}
\textbf{CPD} & \textbf{MMP} & \textbf{FiLM} & & \\
\midrule
 & &  & 0.45 & 0.60 \\
 & & \Checkmark & 0.49 & 0.68 \\
\Checkmark & \Checkmark &  & 0.60 & 0.78 \\
\Checkmark & \Checkmark & \Checkmark & \textbf{0.63} & \textbf{0.83} \\
\bottomrule
\end{tabular}
\end{small}
\end{center}
\vspace{-1em}
\end{table}

\section{Discussion}

Protein molecules are extremely diverse through evolution of 3 billion years \cite{chroma, KEDD}. Since the space of possible protein molecules is much larger than the space of those likely to have human-expected functions~\cite{generative_AI}, it remains a challenge to accomplish controllable protein discovery and optimization. To alleviate this challenge, we introduce \ourapproach in this paper, a CLIP-informed protein editing method via protein-biotext multi-modality learning. Concretely, we execute multi-modality learning of two large language models to effectively align the feature spaces of the protein and biotext following CLIP~\cite{clip}. Building upon the multi-modal pretrained encoders, we adopt an auto-regressive generative decoder to execute cross-modal protein editing, serving as a pioneering work for AI-assisted controllable protein discovery and optimization. 

Extensive experiments are constructed to comprehensively assess the performance of the proposed method. Excitingly, given the text editing instruction, edited protein sequences generated by \ourapproach demonstrate excellent functionality, closely aligning with human-expected functional attributes. We select functional attributes from multiple domains, including enzyme catalytic activity, protein stability and antibody specific binding ability, to execute controllable protein editing towards human-expected functionality. 
Revealed by experimental results, \ourapproach significantly optimizes functional distribution of enzymes (\textit{i.e.}, enzymes initially with poor functionality tend to cluster together with the high-functionality enzymes after being edited by \ourapproach). 
Additionally, \ourapproach successfully designs protein sequences with 16.67\% and 16.90\% stability improvement compared to original protein sequences. And optimized antibodies also form stable, regular 3D structures binding with specific SARS-CoV antigens.
If we take a closer look at the proposed method, the core idea of \ourapproach is to align feature spaces via multi-modality learning and accomplish effective and accurate cross-modal protein editing using features from the protein and biotext domains.

While \ourapproach bears promise in accelerating AI-assisted controllable protein discovery and optimization, there are still areas for improvement and future exploration:
(1) When training the proposed protein editing generator, we did not update any parameters of the pretrained large-scale encoders. It may be beneficial to consider incorporating some Parameter Efficient Fine-Tuning (PEFT) approaches into pretrained encoders in the future.
(2) \ourapproach encodes protein representations primarily on sequences but lacks explicit consideration for structural information. Therefore, incorporating protein structures may lead to more comprehensive understanding of proteins.
(3) The auto-regressive generation manner falls short in designing proteins with a specified sequence length (\textit{i.e.}, auto-regressive generation relies on the probabilistic sampling of the stop token to determine the generated sequence length), leaving some room for exploring more suitable generation paradigm for controllable protein editing. The further exploration will also promote protein discovery and optimization in real-world applications.

\section{Conclusion}
A major unresolved task in biological research, clinical medicine, and biotechnology is achieving controllable protein discovery and optimization. Protein editing approaches offer a potential solution but still rely heavily on laborious high-throughput screening and experimental heuristics.
In this work, we present \ourapproach, a deep learning based method for controllable protein editing and optimization. Following CLIP~\cite{clip}, \ourapproach performs large-scale multi-modal pretraining to align protein and biotext feature spaces and thus well execute the cross-modal protein editing. The target protein we designed exhibited improved functional attributes as expected, and well suited the requirements of multiple domains, including enzyme catalytic activity, protein stability, and antibody specific binding ability.
We hope our work will accelerate the ultimate goal of accomplishing controllable protein discovery and optimization in real-world scenarios.

\bibliographystyle{unsrt}  
\bibliography{ref.bib} 

\clearpage
\appendix

\end{document}